\documentclass{article}

\PassOptionsToPackage{square,numbers}{natbib}
\usepackage[preprint]{neurips_2024}

\usepackage{mathtools}
\usepackage[utf8]{inputenc} 
\usepackage[T1]{fontenc}    
\usepackage{hyperref}       
\usepackage{url}            
\usepackage{booktabs}       
\usepackage{amsfonts}       
\usepackage{nicefrac}       
\usepackage{microtype}      
\usepackage{xcolor}   
\usepackage{amsmath}
\usepackage{amsfonts}
\usepackage{amssymb}
\usepackage{calligra}
\usepackage{calrsfs}
\usepackage[mathcal]{eucal}
\usepackage{mhchem}
\usepackage{tabularx}
\usepackage{booktabs}

\title{Can KANs (re)discover predictive models for Direct-Drive Laser Fusion?}

\author{%
  Rahman Ejaz\thanks{Laboratory for Laser Energetics , University of Rochester} \\
  University of Rochester\\
 \texttt{reja@lle.rochester.edu} \\
  \And
  Varchas Gopalaswamy \\
  University of Rochester \\
  \texttt{vgop@lle.rochester.edu} \\
  \And
  Riccardo Betti \\
  University of Rochester \\
  \texttt{betti@lle.rochester.edu} \\
  \AND
  Aarne Lees \\
  University of Rochester \\
  \texttt{alee.lle.rochester.edu} \\
  \And
  Christopher Kanan \\
  University of Rochester \\
  \texttt{ckanan@cs.rochester.edu} \\
}

\begin{document}

\maketitle

\begin{abstract}
  The domain of laser fusion presents a unique and challenging predictive modeling application landscape for machine learning methods due to high problem complexity and limited training data.  Data-driven approaches utilizing prescribed functional forms, inductive biases and physics-informed learning (PIL) schemes have been successful in the past for achieving desired generalization ability and model interpretation that aligns with physics expectations. In complex multi-physics application domains, however, it is not always obvious how architectural biases or discriminative penalties can be formulated. In this work, focusing on nuclear fusion energy using high powered lasers, we present the use of Kolmogorov-Arnold Networks (KANs) as an alternative to PIL for developing a new type of data-driven predictive model  which is able to achieve high prediction accuracy and physics interpretability. A KAN based model, a MLP with PIL, and a baseline MLP model are compared in generalization ability and interpretation with a domain expert-derived symbolic regression model. Through empirical studies in this high physics complexity domain, we show that KANs can potentially provide benefits when developing predictive models for data-starved physics applications.
\end{abstract}

\section{Introduction}
\label{intro}
 Inertial confinement fusion \citep{nuckolls1972laser} (ICF) is an approach to nuclear fusion power and clean energy that uses high power lasers to achieve conditions similar to stellar interiors in the laboratory, and is governed by complicated and deeply nonlinear underlying physics. Sophisticated multi-physics solvers are traditionally used to understand the physics of ICF, but despite significant progress in high-fidelity integrated simulation capabilities for ICF \citep{igumenshchev2010crossed,igumenshchev2017three}, simulations alone are not sufficient to predict outcomes of experiments \textit{a priori} and their utility to design future experiments remains limited. This lack of predictive capability is a major impediment towards realizing nuclear fusion via inertial confinement as a viable clean energy source.
 
 \par To bridge the gap between simulation predictions and experimental outcomes, data-driven approaches are utilized. Past approaches have relied on using physics intuition for prescribing functional forms (piece-wise power laws) for physically motivated models using quantities derived from simulations \cite{gopalaswamy2019tripled,lees2021experimentally,lees2023understanding}. In order to relax the hypothesized functional form assumptions similar models to \cite{gopalaswamy2019tripled,lees2021experimentally,lees2023understanding} were developed in \cite{ejaz2024deep}, where multi-layer perceptrons (MLPs) were used due to their function approximation abilities indicated by the universal approximation theorem. However, \cite{ejaz2024deep} found that for their small-sized dataset the inclusion of applicable inductive biases was needed to improve prediction generalization and model interpretation to be in accordance with physics expectations. 
 
 \par Our work presented attempts to address physics-based regression problems that require minimal assumptions on the underlying functional terms to explain the data, but are not amenable to physics-informed or augmented solutions\cite{liu2021physics,takbiri2024physics,davini2021using}. This can be either due to complexity in explicitly formulating desired model properties into discriminative penalties as is conventionally done in physics-informed learning (PIL) and inapplicability of model design methods such as physics-augmented learning~\cite{liu2021physics}. In essence we showcase our work to an audience where limited data makes learning desired model properties solely from data alone challenging and it is not efficient or obvious on how to effectively construct a $\mathcal{L}_{phys}$ term (defined in Eq.~\ref{eq.1}) which is augmented with the usual loss functions,
 \begin{equation}
    \mathrm{\mathcal{L}_{phys} =  \sum_{i}^{\textit{n}}\gamma_{i}\Psi_{i}({f(\textit{x}_{i};\theta)}) }\label{eq.1}
\end{equation}

where \textit{n} is the input feature dimension, $\mathrm{\Psi}$ is an operator which captures deviations from the desired property, $\mathrm{f(\textbf{x}};\theta)$ is the neural network and $\gamma$ is the strength of the penalty. Constructing an effective ${\mathcal{L}_{phys}}$ is a typical requirement for multiple methods that interleave physics with deep learning. 
 
\par We show that in our particular use case of developing a new type of predictive model for direct-drive laser fusion the Kolmogorov-Arnold Networks (KANs) proposed by \cite{liu2024kan} has shown modest improvement for generalization on out-of-distribution (ood) data and comes closest in its interpretation with a domain expert derived model when compared with applicable PIL schemes. The PIL schemes are implemented through the $\mathcal{L}_{phys}$ term and are constructed to include all the $\mathrm{\Psi}$ that can appropriately be considered prior knowledge. The ICF prediction model presented uses an input parameterization that makes it more general in directly relating the prediction quantity with experimental design features compared with the works of \cite{gopalaswamy2019tripled,lees2021experimentally}, thus aiding in experimental design. We hypothesize that the learnable activation functions feature of KANs acts as a form of conformal/learnable inductive bias through learning of internal degrees of freedom (B-splines)\cite{liu2024kan}. The precise nature of the external and internal degrees of freedom and their role in making KANs successfully learn details of the physical system will be explored future works.

\section{Dataset and Modeling Task}
The empirical data used in this study come from cryogenic implosion experiments conducted on the OMEGA Laser Facility from 2016 to present that are relevant to fusion energy. The substantial difficulty in conducting these experiments limits the data to $\sim$300 samples, which precludes the use of many successful ML techniques. Using this database, the objective is to develop a model of the experimental fusion yield, which is the number of reactions and therefore is proportional to the energy released by nuclear fusion.
The fusion yield obtained in these experiments depends on the energy delivered to the fusion fuel target, details of energy delivery to the target via the laser power history\cite{nuckolls1972laser}, specifications of the fusion fuel payload\cite{nuckolls1972laser}, and 3-dimensional effects \cite{lees2021experimentally,lees2023understanding}. Using a parameterization of the experimental fusion yield dependencies, the modeling task is to learn a mapping function which is given in general terms below,
\begin{equation}
    \mathrm{{Y^{exp} = \mathrm{f}(\textit{E}_{L},R_{out},\hat{\textit{M}},\hat{\textit{R}},\frac{R_{b}}{R_{t}},\frac{\alpha}{\textit{IFAR}},\textit{CR},\hat{\textit{V}},\hat{\textit{Y}},\textit{T}_{\frac{max}{min}},YOC_{He}^{sim}}}) 
    \label{eq:3}
\end{equation}
where the inputs are described in more detail in Appendix~\ref{appendix_a1} and are physically motivated - however, the precise details of how the experimental yield depends on these inputs and their interactions is not yet well understood. We note that such fusion yield dependence parameterizations have been successful in the works of \cite{gopalaswamy2019tripled,lees2021experimentally,lees2023understanding}, though the parameterization in Eq~\ref{eq:3} is more general than the previous mentioned works.

\section{Methods}
\label{methods}
\vspace{-0.25cm}
\paragraph{Evaluation Criterion}
The KAN based mapping model is compared in its prediction ability on ood experimental samples and in its agreement with the expert derived mapping function given in Eq.~\ref{eq:domainexpert}. It should be noted that the domain expert derived model is not considered the "oracle" mapping relation as simplifying assumptions such as the absence of interactions between the input parameters is assumed in the formulation. The presence of complex interactions due to physics considerations is hypothesized \cite{lees2023understanding,ejaz2024deep}, however the nature and extent of the interactions is unknown. This motivates the use of automatic functional "finding" methods. However, when using these methods the inferred dependencies of $\mathrm{Y^{exp}}$ on input parameters (shown in Eq.~\ref{eq:3}) found 
need to be on average consistent with the physics of the model in Eq.~\ref{eq:domainexpert}. This is also used as an evaluation criterion and informs which model is preferable. 
\par
Due to design intent by physicists at OMEGA, the empirical database is curated into sets called "campaigns". However, it is not obvious how distinctions between campaigns correspond to distinctions in input parameter space. Therefore, for ood tests k-means clustering \cite{jain1988algorithms} is utilized to partition the dataset. The number of clusters is determined using a silhouette method \cite{rousseeuw1987silhouettes} with the scan being restricted to the total number of unique campaigns in the dataset (29 campaigns). Given the clusters the dataset becomes $\mathcal{D} = \{ \mathcal{D}^{1},\mathcal{D}^{2},...,\mathcal{D}^{N} \}$ where $\mathbf{N}$ is the total number of clusters ($\textit{N}=6$). A test dataset $\mathcal{D}^{\textit{test}} \in \mathcal{D}$ is chosen and correspondingly the train dataset becomes $\mathcal{D}^{train} = \{ \mathbf{X} | \mathbf{X} \in \mathcal{D} \land \mathbf{X} \notin \mathcal{D}^{test} \}$ where $\mathbf{X}$ are the input-target pairs. The mean squared error (MSE), $\mathrm{\frac{1}{\text{N}}\sum_{\text{i}=1}^{\text{N}} (\mathbf{\hat y}_{i}-\mathbf{y}_{i})^{2}}$, where $\mathbf{\hat y}$ is the model prediction and $\mathbf{y}$ is the ground truth is compared across the models for all clusters used as $\mathcal{D}^{test}$. The prediction error $\mathrm{{\frac{\mathbf{\hat y}_{i}-\mathbf{y}_{i}}{\mathbf{y}_{i}}}} \times 100$ ($<> \pm \sigma$) on a new subset of experiments is also compared using these models as a means to probe prediction ability for future experiments.

\subsection{Domain Expert Model}
Intuition and physics knowledge is used to derive the domain expert model which is an evolution of the models in ~\cite{gopalaswamy2019tripled,lees2021experimentally,lees2023understanding}. This model is represented as a piece-wise power law and is given below in Eq.~\ref{eq:domainexpert}.
\begin{equation}
    \resizebox{1.03\hsize}{!}{$\mathrm{
    Y^{\text{exp}} = E_{L}^{2.3}R_{\text{out}}^{-2.6}\hat{M}^{2.9}\hat{\textit{R}}^{26.5}\underset{\frac{R_{b}}{R_{t}}<0.86}{\left(\frac{R_{b}}{R_{t}}\right)^{1.6}}\underset{0.86<\frac{R_{b}}{R_{t}}<1}{\left(\frac{R_{b}}{R_{t}}\right)^{-3.0}}\underset{\frac{R_{b}}{R_{t}}>1}{\left(\frac{R_{b}}{R_{t}}\right)^{0}}\underset{<1}{\left(\frac{\alpha}{\text{IFAR}}\right)^{0.45}}\underset{>1}{\left(\frac{\alpha}{\text{IFAR}}\right)^{-0.1}}\hat{\text{V}}^{2.02}\hat{\text{Y}}^{0.78}T_{\frac{\text{max}}{\text{min}}}^{(-1.32)}(YOC_{\text{He}}^{\text{sim}})^{1.26}
    }
    $
    \label{eq:domainexpert}
    }
\end{equation}
One major drawback of this parametrization strategy is the lack of interaction effects between the terms. Simple physics intuition, verified in some cases by multi-physics simulations\cite{gopalaswamy2021using} show that interaction effects can be important, but it is non-trivial to formulate them correctly in a closed-form parametrization. Capturing these effects is of high importance, not only to better predict the outcomes of future designs, but also because the details of interactions can point towards physical behaviors that are important in ICF experiments. 

\subsection{Physics-informed Model}
For the MLP PIL model, a fully connected MLP, $\mathbf{\textit{f}_{MLP}}(\textbf{X}) = (\sigma\mathbf{W_{N}} \circ \sigma\mathbf{W_{N-1} } \circ \dots \circ \sigma\mathbf{W_{1}}) \textbf{X} $  with a loss function $\mathcal{L} = \mathcal{L}_{MSE} + \mathcal{L}_{physics}$ is used. The $ \mathcal{L}_{physics}$ is constructed such that it enforces properties that are definitively considered prior knowledge (explained in Appendix ~\ref{appendix_a3}).
\begin{equation}
    \mathrm{\mathcal{L}_{physics} =\gamma_{1} \frac{\hat\partial {f(\mathbf{x};\theta)}}{\partial \textit{T}_{\frac{max}{min}} } + \gamma_{2} \frac{\bar\partial{f(\mathbf{x};\theta)}}{\partial YOC_{He}^{sim} }} \label{eq:4}
\end{equation}
Where, $\frac{\hat{\partial} f(\mathbf{x};\theta)}{\partial T_{\frac{\text{max}}{\text{min}}}} = 
\begin{cases} 
0 & \text{if } \frac{\partial f(\mathbf{x};\theta)}{\partial T_{\frac{\text{max}}{\text{min}}}} \leq 0 \\
\frac{\partial f(\mathbf{x};\theta)}{\partial T_{\frac{\text{max}}{\text{min}}}} & \text{if } \frac{\partial f(\mathbf{x};\theta)}{\partial T_{\frac{\text{max}}{\text{min}}}} > 0 
\end{cases}$ and $\frac{\bar{\partial} f(\mathbf{x};\theta)}{\partial YOC_{He}^{\text{sim}}} = 
\begin{cases}
\frac{\partial f(\mathbf{x};\theta)}{\partial YOC_{He}^{\text{sim}}} & \text{if } \frac{\partial f(\mathbf{x};\theta)}{\partial YOC_{He}^{\text{sim}}} < 0 \\
0 & \text{if } \frac{\partial f(\mathbf{x};\theta)}{\partial YOC_{He}^{\text{sim}}} \geq 0 
\end{cases}$

For the baseline MLP model only the $\mathcal{L}_{MSE}$ loss is used while the rest of the training configuration and model architecture is kept the same as the MLP with PIL model.

\subsection{KAN Model}
Differing from MLPs, KANs have no linear learnable weights  $\mathbf{{f}_{KAN}}(\textbf{X}) = (\mathbf{\Phi_{N}} \circ \mathbf{\Phi_{N-1} } \circ \dots \circ \mathbf{\Phi_{1}}) \textbf{X} $, each weight parameter in $\Phi$ is replaced by a learnable univariate function constructed using B-spline basis functions \cite{liu2024kan}. The KAN based model is trained using solely the $\mathcal{L}_{MSE}$ loss.

\section{Results and Discussion}
\label{results}

\begin{figure}[ht]
    \centering

    \begin{minipage}{0.24\textwidth}
        \centering
        \includegraphics[width=\textwidth]{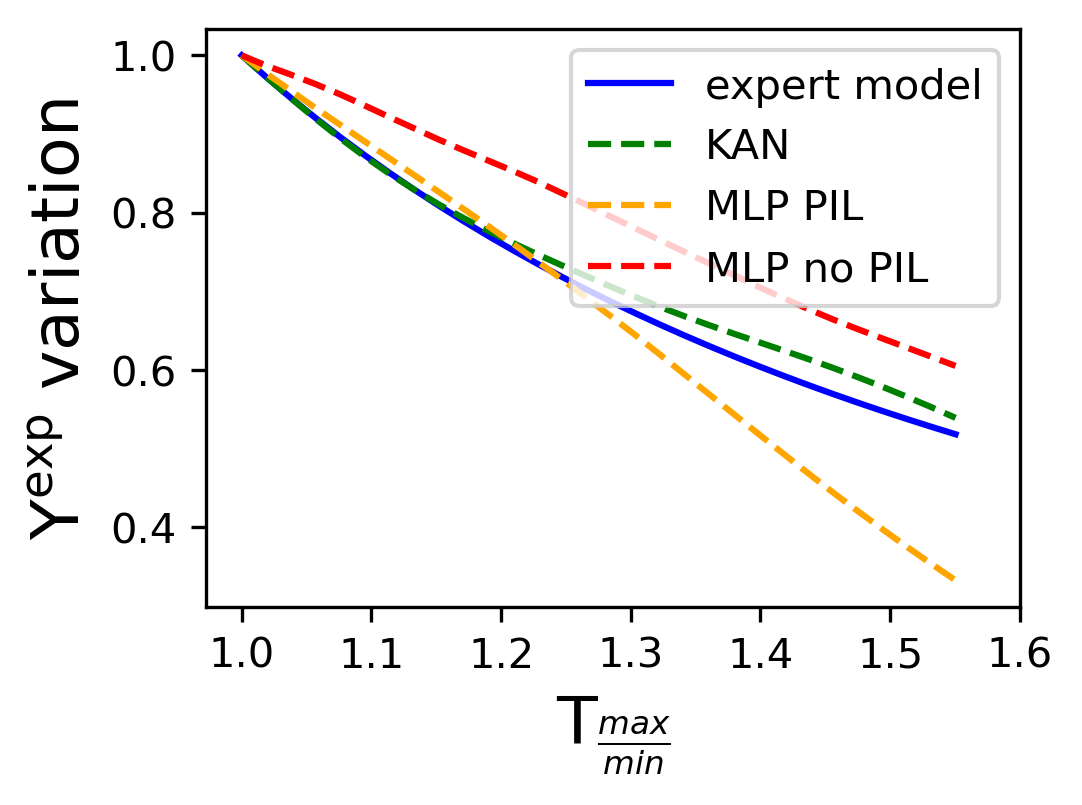}
    \end{minipage}\hfill
    \begin{minipage}{0.24\textwidth}
        \centering
        \includegraphics[width=\textwidth]{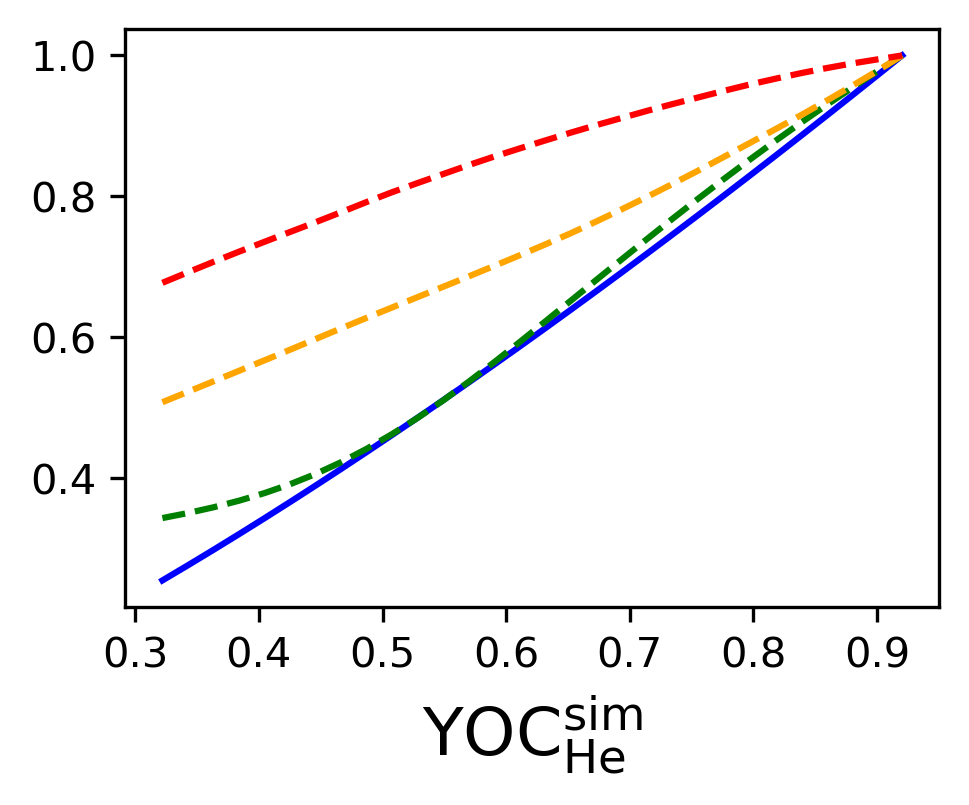}
    \end{minipage}\hfill
    \begin{minipage}{0.24\textwidth}
        \centering
        \includegraphics[width=\textwidth]{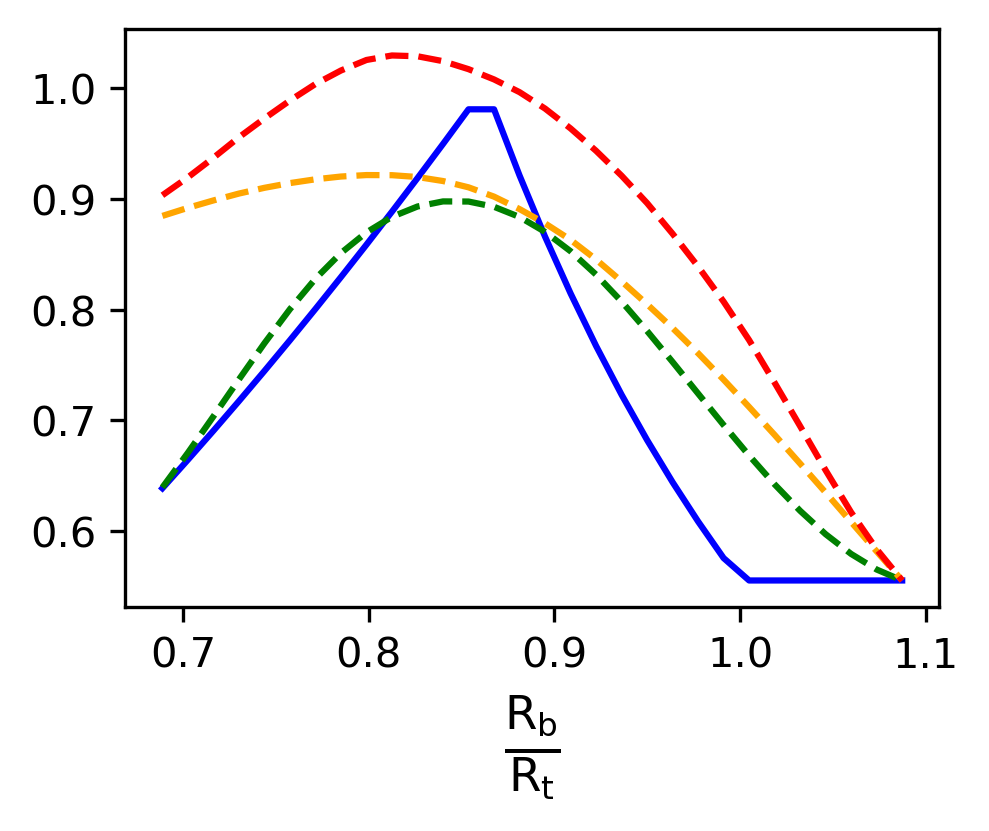}
    \end{minipage}\hfill
    \begin{minipage}{0.24\textwidth}
        \centering
        \includegraphics[width=\textwidth]{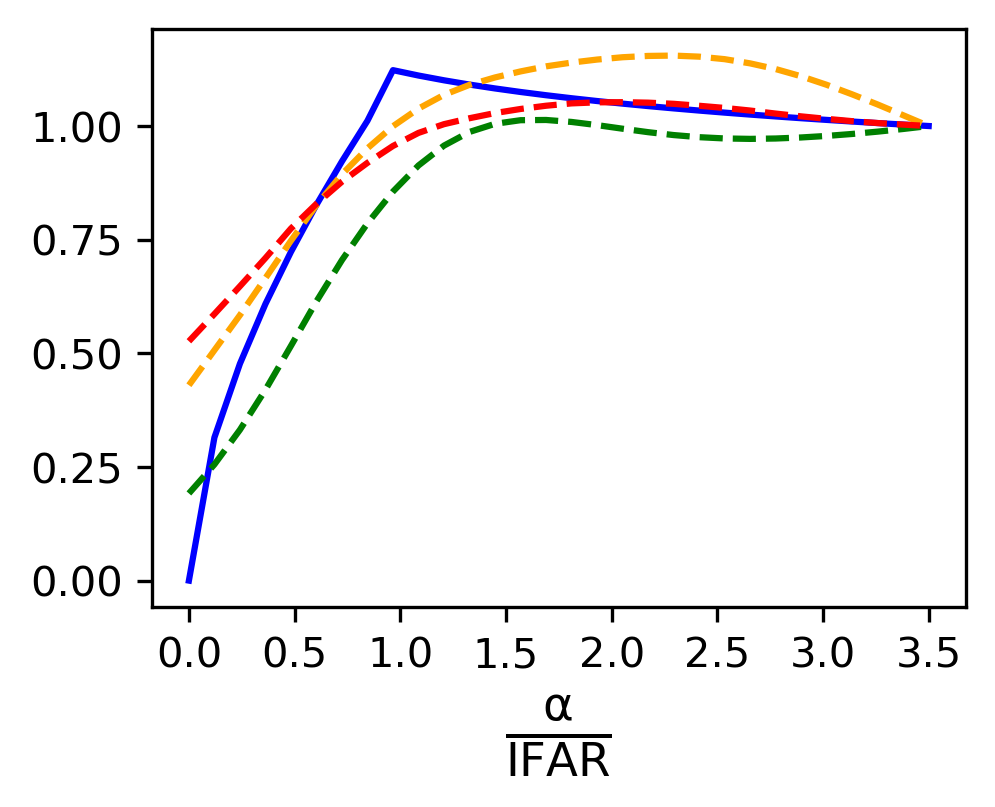}
    \end{minipage}

    \vspace{0.5cm}
    \begin{minipage}{0.24\textwidth}
        \centering
        \includegraphics[width=\textwidth]{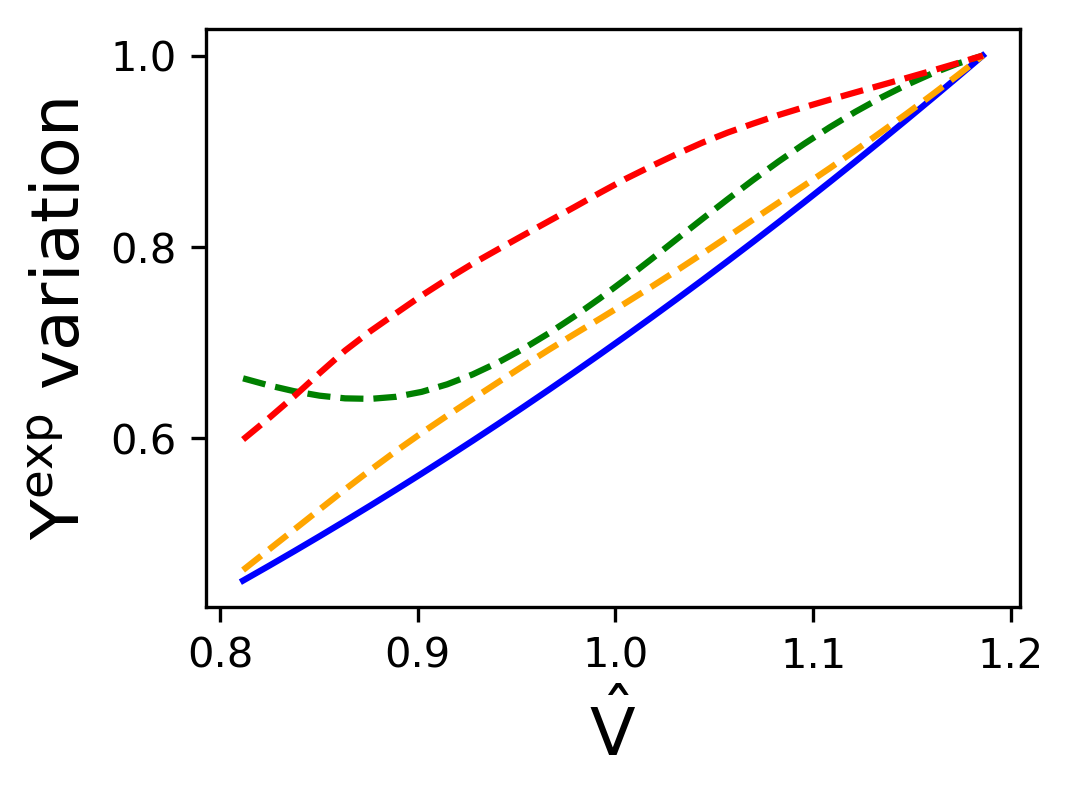}
    \end{minipage}\hfill
    \begin{minipage}{0.24\textwidth}
        \centering
        \includegraphics[width=\textwidth]{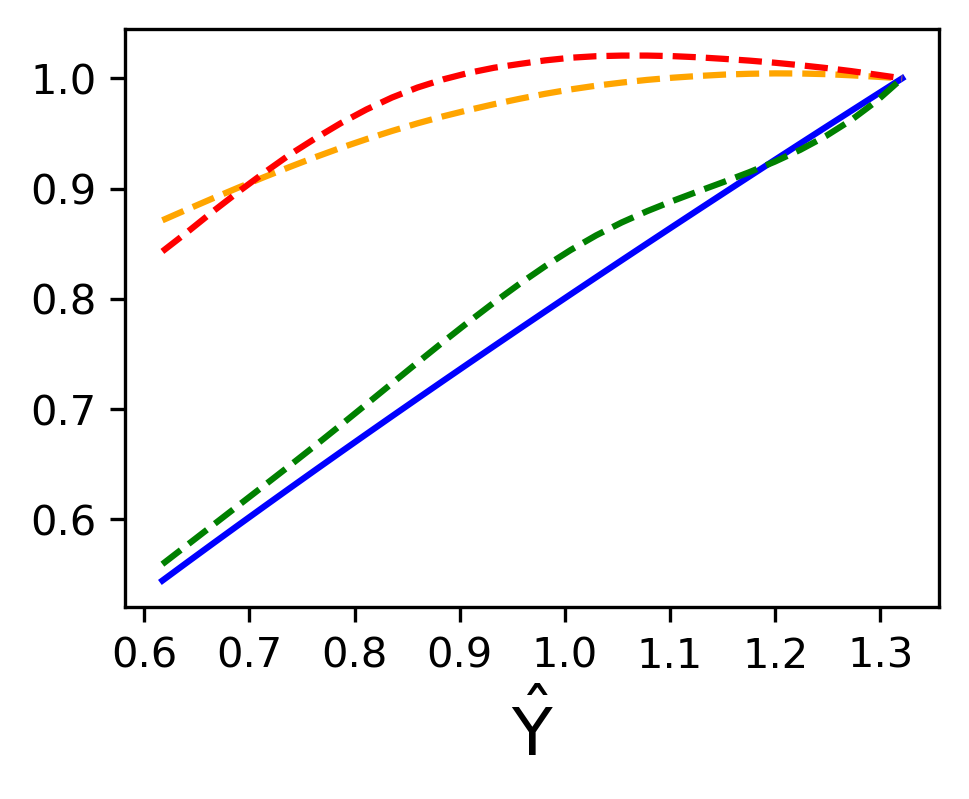}
    \end{minipage}\hfill
    \begin{minipage}{0.24\textwidth}
        \centering
        \includegraphics[width=\textwidth]{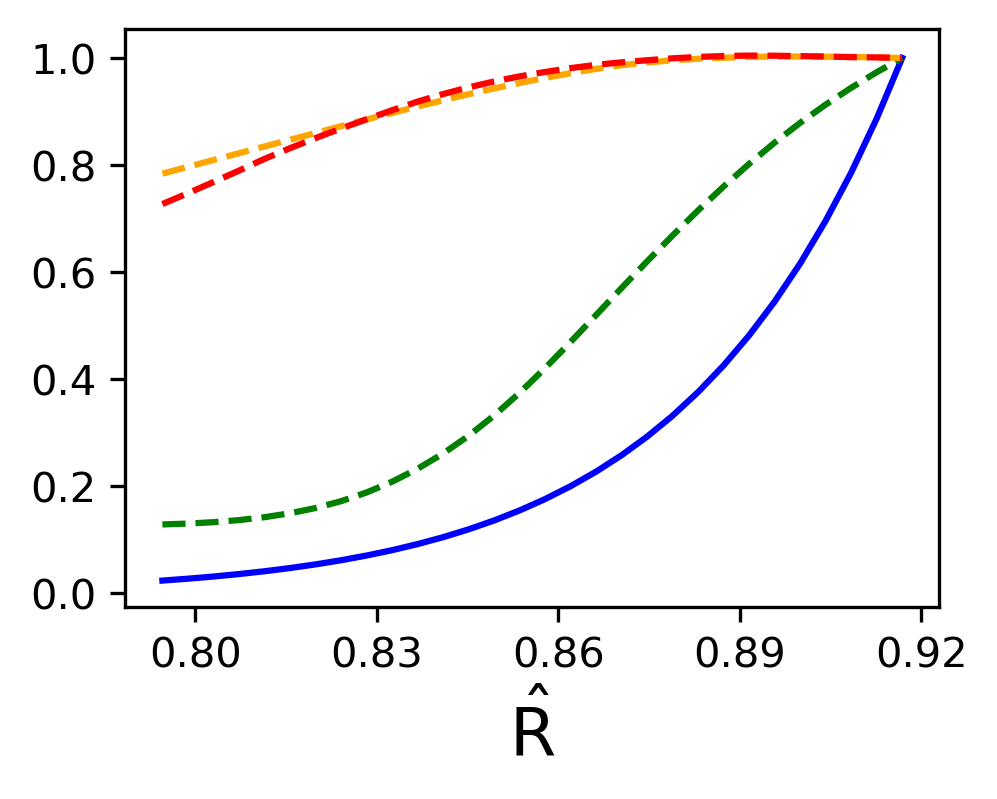}
    \end{minipage}\hfill
    \begin{minipage}{0.24\textwidth}
        \centering
        \includegraphics[width=\textwidth]{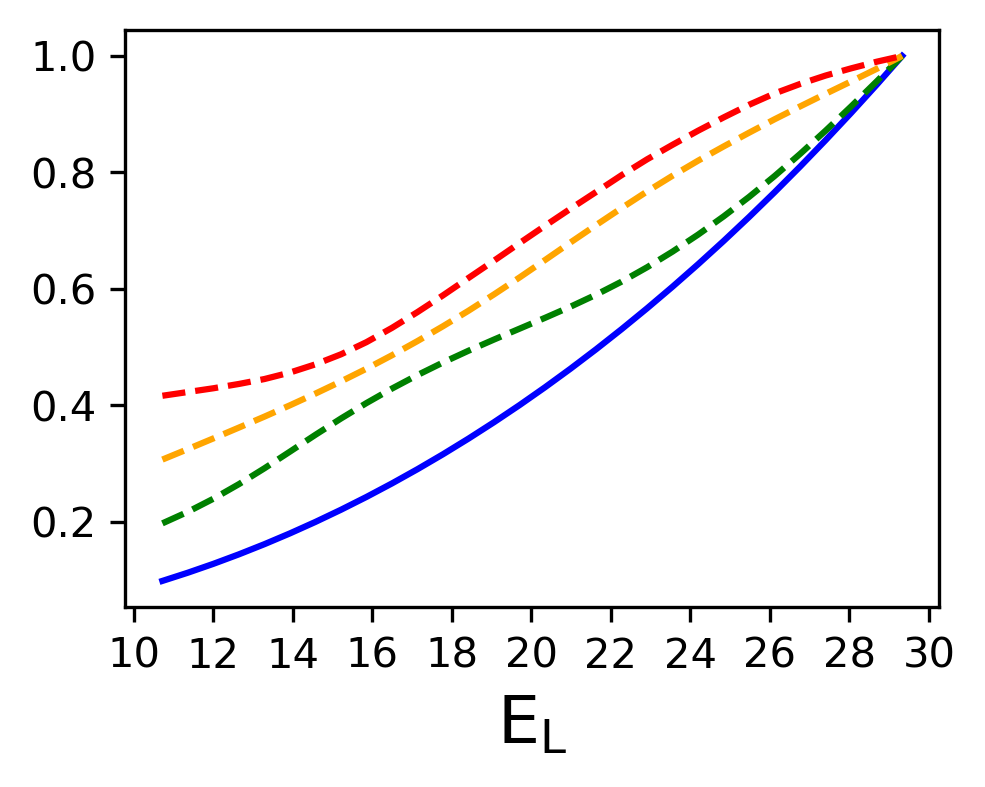}
    \end{minipage}

    \caption{Inferred $\mathrm{Y^{exp}}$ variation across different models compared with the expert derived model, shown in arbitrary units (A.U)}
    \label{fig:yield_dependencies}
\end{figure}
\paragraph{Main results} To interpret the learned behavior of the model, the partial dependence method \cite{friedman2001greedy} is utilized where $\textit{f(\textbf{x}};\theta)$ is varied as a function of its input arguments. The average $\mathrm{Y^{exp}}$ variation on input parameters from the models is shown in Fig.~\ref{fig:yield_dependencies} along with the domain expert derived model dependencies for direct comparison. All the models generally align with the expert derived model. However, the KAN model is in better agreement with the expert model as illustrated in Fig.~\ref{fig:yield_dependencies}. Important notable differences between the models are for the $\frac{R_{b}}{R_{t}}$ and $\mathrm{\frac{\alpha}{\textit{IFAR}}}$ terms. Due to multiple competing physics effects\cite{lees2023understanding,gopalaswamy2021using} the optimum value of $\frac{R_{b}}{R_{t}}$ $\approx 0.85$. While both models capture this nature, the strong dependence of $\mathrm{Y^{exp}}$ with $\frac{R_{b}}{R_{t}}$ for $\frac{R_{b}}{R_{t}}$ $< 0.85$ is more accurately captured by the KAN model. The rapid fall off of for $\mathrm{Y^{exp}}$ below 0.85 is justified in the expert model based on detailed physics studies\cite{gopalaswamy2021using} and only the KAN model is able to correctly recover this behavior. 
We can make a similar case for the $\mathrm{\frac{\alpha}{\textit{IFAR}}}$ term, though in this case the justification for the dependence inferred by the expert model is based largely on physical intuition. As $\mathrm{\frac{\alpha}{\textit{IFAR}}}$ decreases, $\mathrm{Y^{exp}}$ is expected to undergo a transition from weak dependence to strong dependence, at which point $\mathrm{Y^{exp}}$ is expected to drop dramatically. KAN also recovers this behavior with more fidelity than the MLPs making it the preferable choice. Although, it can be argued that a suitable $\mathcal{L}_{physics}$ to capture the above mentioned effects in the PIL model can be devised, these aspects of $\frac{R_{b}}{R_{t}}$ and $\mathrm{\frac{\alpha}{\textit{IFAR}}}$ have been uncovered through detailed and expensive high-fidelity physics simulations \cite{gopalaswamy2021using,goncharov2014improving} and dedicated experiments and as a result are included in the domain expert model. Thus, we postulate that information of this type isn't available \textit{a priori} and shouldn't be included when investigating the merits of PIL schemes. 
\vspace{-0.25cm}
\paragraph{Out-of-distribution tests} The KAN model on average ($<> \pm \sigma$) achieves an incremental advantage on MSE scores across the clusters ($\text{0.00394} \pm \text{0.00410}$ vs next best $0.00926 \pm 0.00506$). This translates to prediction errors of $(\text{0.6}\pm\text{5.9})\%$ using the KAN model, $(\text{1.2}\pm\text{9.5})\%$ using MLP PIL model and $(\text{9.6}\pm\text{4.2})\%$ using the MLP model when tasked to predict the outcomes of new experiments. The prediction error using the domain expert model is $(\text{12.8}\pm\text{10.4})\%$.
\vspace{-0.25cm}
\paragraph{Discussion} The benefit of the KAN model in terms of generalization ability is not conclusive solely from these ood tests. The new experiments mentioned are not true ood samples as they have overlap with previous experiments. This is often the case in the domain of laser direct-drive fusion. Furthermore, the data comes from real-world experiments and thus has random variations which also complicates any benefit interpretation. To further elucidate any generalization benefit the use of variational autoencoders based segmentation\cite{yu2020auto} of the data would be investigated in future works. We note that for the $\mathrm{\textit{T}_{\frac{max}{min}}}$ and $\mathrm{YOC_{He}^{sim}}$ terms the MLP PIL model can be brought in closer agreement with the expert model by tuning the hyperparameters $\mathrm{\gamma_{1}}$ and $\mathrm{\gamma_{2}}$. An added advantage of KANs is that for instances of nonphysical behavior such as the curl up behavior for low argument values in the $\mathrm{{YOC}_{He}^{sim}}$ and $\hat{\textit{V}}$ plots, the grid extension technique~\cite{liu2024kan} could be performed post-training to fine-tune the spline hyperparameters which can mitigate such nonphysical behaviors. However, we leave this investigation on the choice of spline/grid hyperparameters pre-training and post-training for future works. In summary, the KAN model comes closest to reproducing the aggregate behavior of an expert derived model while also allowing for interaction effects between the inputs. The physical interpretation and validity of these interaction effects will be investigated as the KAN is used to design new experiments, and will be discussed in subsequent publications.

\begin{ack}
This material is based upon work supported by the Department of Energy under Award Nos. DE-SC0024381, DESC0022132, DE-SC0021072 and DE-SC0024456 as well as the Department of Energy (National Nuclear Security Administration) University of Rochester “National Inertial Confinement Program” under Award No. DE-NA0004144. This report was prepared as an account of work sponsored by an agency of the United States Government. Neither the United States Government nor any agency thereof, nor any of their employees, makes any warranty, express or implied, or assumes any legal liability or responsibility for the accuracy, completeness, or usefulness of any information, apparatus, product, or process disclosed, or represents that its use
would not infringe privately owned rights. Reference herein to any specific commercial product, process, or service by trade name, trademark, manufacturer, or otherwise does not necessarily constitute or imply its endorsement, recommendation, or favoring by the United States Government or any agency thereof. The views and opinions of authors expressed herein do not necessarily state or reflect those of the United States Government or any agency thereof.
\end{ack}

\bibliography{references}


\newpage
\appendix

\section{Appendix / supplemental material}
\subsection{Input Parameter Description}
\label{appendix_a1}
\begin{table}[h]
    \centering
    \caption{Model input parameters}
    \begin{tabular}{|l|l|}
        \hline
        \textbf{Parameter} & \textbf{Description} \\
        \hline
        $\textit{E}_{L}$ & Total laser energy on fusion fuel target.\\
        $R_{out}$ & Target outer radius.\\
        $\hat{\textit{M}}$ & $\frac{\textit{Mass}_{\textit{fuel in ice form}}}{\textit{Mass}_{\textit{target total}}}$\\
        $\hat{\textit{R}}$ &  $\frac{\textit{Radius}_{\textit{inner}}}{\textit{Radius}_{\textit{outer}}}$\\
        $\frac{R_{b}}{R_{t}}$ & Captures effects of $\sigma_{rms}$ illumination nonuniformity\cite{lees2023understanding,gopalaswamy2021using}.\\
        $\frac{\alpha}{\textit{IFAR}}$ & Stability parameter\cite{lees2023understanding,goncharov2014improving}, captures effect of short-wavelength perturbations.\\
        $\textit{CR}$ & Initial outer shell radius/inner shell radius at maximum compression.\\
        $\hat{\textit{V}}$ & Ad-hoc parameter capturing nonlinear effect of pulse shape on maximum achieved velocity.\\
        $\hat{\textit{Y}}$ & Ad-hoc parameter capturing nonlinear effect of pulse shape on simulated yield.\\
        $\textit{T}_{\frac{max}{min}}$ & Captures degradation magnitude when ion temperature asymmetries are present.\\
        $\mathrm{YOC_{He}^{sim}}$ & Captures the degrading effect of $\mathrm{{^3He}}$ gas contamination due to fuel target's age.\\
        \hline
    \end{tabular}
    \label{tab:example_table}
\end{table}

\subsection{Model architecture, hyperparameters and training details}
\label{appendix_a2}

\paragraph{Architecture and hyperparameter optimization}
An informal grid search was used to optimize the model architecture and hyperparameter configurations. In future works a more formal approach for hyperparameter optimization and architecture ablation would be used utilizing the bayesian hyperprop strategy \cite{wu2019hyperparameter}.
\vspace{-0.25cm}
\paragraph{Compute}
All models are trained on a desktop workstation with a single Nvidia RTX 6000 GPU. The KAN model takes $\approx 30$ mins to train. The MLP and MLP PIL models take $\approx 180$ mins to train on a single train-test split.
\paragraph{Train-test split} Scikit-learn\cite{scikit-learn} KFold was using for 21 train-test partitions, all results apart from out-of-distribution results are presented using split number 8.

\paragraph{K-means for out-of-distribution tests} Scikit-learn\cite{scikit-learn} silhouette\_score and KMeans were used to obtain clusters for out-of-distribution tests, six clusters were used.

\begin{table}[h]
\centering
\caption{\label{tab:table1}MLP PIL and MLP configuration}
\begin{tabular}{cc}
\textbf{Hyperparameter}& \textbf{Value}  \\ \hline
Input layer width & 16 (extra parameters for one-hot encoding of target composition)\\
Fully connected hidden layer widths & 71-71-71\\
Learning rate & 1e-3  \\
Drop-out regularization & 0.1 \\
Optimizer & Adam~\cite{kingma2014adam}\\ 
Batch size & 11\\
Epochs & 3000\\
Early stopping on & train loss\\
Loss function MLP & $\mathcal{L}_{MSE}$ \\
Loss function MLP PIL &  $\mathcal{L}_{MSE}$ + $\mathcal{L}_{phys}$\\
\hline
\end{tabular}
\end{table}

\begin{table}[h]
\centering
\caption{\label{tab:table2} KAN training configuration. The implementation is utilized from the following source,
{\texttt https://github.com/Blealtan/efficient-kan.git}}
\begin{tabular}{cc}
\textbf{Hyperparameter}& \textbf{Value}  \\ \hline
Input layer width & 16 (extra parameters for one-hot encoding of target composition)\\
Number of layers & 7\\
Layer widths & 71 \\
Grid size & 5\\
Spline order & 3\\
Learning rate & 1e-3  \\
Drop-out regularization & 0.2 \\
Optimizer & Adam~\cite{kingma2014adam}\\ 
Batch size & 11\\
Epochs & 500\\
Early stopping on & train loss\\
Loss function & $\mathcal{L}_{MSE}$\\
\hline
\end{tabular}
\end{table}

\newpage
\subsection{$\mathcal{L}_{physics}$ explanation}
\label{appendix_a3}

The $\mathcal{L}_{physics}$ in Eq.~\ref{eq:4} for both terms can naively be applied as a penalty whenever deviations from the MLP PIL model is exhibited during training as for both terms simple physics consideration can be used to establish the validity of the expected trend.

\par The $\textit{T}_{\frac{max}{min}}$ can be trivially shown\cite{woo2020inferring} to be related to the residual kinetic energy of the fusing plasma. Ideally, all the kinetic energy is converted to internal energy of the fusing plasma, which then results in fusion. Reducing the amount of kinetic energy $\to$ internal energy monotonically reduces the reaction rate and therefore the number of reactions, thereby requiring that $\mathrm{\frac{\partial\mathrm{Y^{exp}}}{\partial\textit{T}_{\frac{max}{min}}}}$ must be negative.

\par The term $YOC_{He}^{sim}$ expresses the degrading effect resulting due to the target's age which is the time between when target assembly begins and when the experiment is executed. As the target age increases $YOC_{He}^{sim}$ $\to$ 0. Due to natural $\beta$ decay some of the fusion fuel is constantly converted in $\mathrm{{^3He}}$ which through established physics causes increased radiation losses and because this gas accumulates in the gas region of the fusion fuel target this leads to a reduction in the amount of fuel compression that can achieved in the laser fusion scheme. It can simply be deduced that both of these effects monotonically reduce $\mathrm{Y^{exp}}$.

\par In comparison to these two effects, the dependence of $\mathrm{Y^{exp}}$ on $\frac{R_{b}}{R_{t}}$ or $\mathrm{\frac{\alpha}{\textit{IFAR}}}$ is not trivial. Both these terms have competing effects that either reduce or increase $\mathrm{Y^{exp}}$, $\frac{R_{b}}{R_{t}}$ going down means more degradation seed\cite{gopalaswamy2021using}, but also means better laser inefficiency mitigation = better energy transfer. Not obvious which one is stronger, where the trade-off is maximized, if its uni-modal, monotonic, etc, $\mathrm{\frac{\alpha}{\textit{IFAR}}}$ going down means better simulation performance but worse experiment performance. Again, not clear if its uni-modal, bimodal, monotonic, etc. It needs to be learned from the data.


\end{document}